\documentclass[conference]{IEEEtran}
\IEEEoverridecommandlockouts
\usepackage{cite}
\usepackage{amsmath,amssymb,amsfonts}
\usepackage{algorithmic}
\usepackage{graphicx}
\usepackage{textcomp}
\usepackage{xcolor}
\usepackage{tabularx}
\usepackage{array} % for better column alignment
\usepackage{booktabs}
\usepackage{amssymb} % For \dag

\def\BibTeX{{\rm B\kern-.05em{\sc i\kern-.025em b}\kern-.08em
    T\kern-.1667em\lower.7ex\hbox{E}\kern-.125emX}}
\begin{document}

\title{Generative Models and Connected and Automated Vehicles: A Survey in Exploring the Intersection of Transportation and AI}

\author{\IEEEauthorblockN{Bo Shu}
\IEEEauthorblockA{\textit{School of Engineering Science} \\
\textit{Shandong Xiehe University}\\
Jinan, 250107, PR China \\
shubo@sdxiehe.edu.cn}
\and
\IEEEauthorblockN{Yiting Zhang}
\IEEEauthorblockA{\textit{Department of Electrical Engineering} \\
\textit{Northwestern University}\\
Evanston, IL \\
yitingzhang2025@u.northwestern.edu}
\and
\IEEEauthorblockN{Saisai Hu}
\IEEEauthorblockA{\textit{Department of Computer Science} \\
\textit{Pace University}\\
New York, NY \\
SH50061N@pace.edu}\\
\and
% --- Dummy block for 2nd row, left ---
\IEEEauthorblockN{~~~~~~~~~~~~~~~~~~~~~~~~~~~~}
\IEEEauthorblockA{~}
\and
% --- 4th Author, centered on 2nd row ---
\IEEEauthorblockN{Dong Shu$^{*}$\thanks{$^{*}$ Corresponding Author}}
\IEEEauthorblockA{\textit{Department of Computer Science} \\
\textit{Northwestern University}\\
Evanston, IL \\
dongshu2024@u.northwestern.edu}
\and
% --- Dummy block for 2nd row, right ---
\IEEEauthorblockN{~}
\IEEEauthorblockA{~}
}

\maketitle

\begin{abstract}
This report investigates the history and impact of Generative Models and Connected and Automated Vehicles (CAVs), two groundbreaking forces pushing progress in technology and transportation. By focusing on the application of generative models within the context of CAVs, the study aims to unravel how this integration could enhance predictive modeling, simulation accuracy, and decision-making processes in autonomous vehicles. This thesis discusses the benefits and challenges of integrating generative models and CAV technology in transportation. It aims to highlight the progress made, the remaining obstacles, and the potential for advancements in safety and innovation.

\end{abstract}

\section{Introduction}
In the rapidly evolving landscape of technology, two fields have emerged as frontrunners in shaping the future of our society: Generative Models in artificial intelligence (AI) and Connected and Automated Vehicles (CAVs) \cite{b54}. Generative Models, a cornerstone of AI, are algorithms designed to generate response similar to, but distinct from, data they have been trained on, enabling applications ranging from image and text generation to complex simulations \cite{b55}. Connected and Automated Vehicles, on the other hand, represent the advancement in transportation, merging connectivity, automation, and intelligence to enhance safety, efficiency, and the driving experience. 

% TODO 简单介绍几个intersection的例子
The intersection of these two groundbreaking technologies offers a promising avenue for research and innovation \cite{b56}. By merging the importance of Generative Models in transforming content creation and decision-making processes with CAVs approach to mobility, logistics, and urban planning, researchers have tapped into new potentials in vehicle intelligence, simulation accuracy, and decision-making capabilities. This synergy could lead to more sophisticated predictive models for vehicle behavior, enhanced safety features through realistic simulation environments, and even innovations in vehicle design and traffic management systems. 

% challange
Despite the previous mentioned success, there are still several challenges on the fields remain unsolved. One of the pivotal challenges faced by CAVs and Generative Models revolves around the integration of these technologies in real-world applications, particularly concerning safety and reliability \cite{b57}. For CAVs, ensuring safety in unpredictable traffic conditions and diverse environments remains a significant obstacle. The vehicles must interpret complex scenarios and make split-second decisions, a challenge compounded by the current limitations in AI's ability to fully understand nuanced human behaviors and unforeseen circumstances \cite{b58}. Generative Models, on the other hand, face issues of data privacy and decision reliability. These challenges threaten both the input and output of models. Users are afraid to provide models with all their data, and they can't fully trust the generated output \cite{b59}. Fixing those challenges require advancements in AI's understanding of the physical world and its ability to generate data that faithfully represents it, ensuring that CAVs can operate safely and effectively in any given situation.

This survey aims to delve into the challenges and relationship between Generative Models and Connected and Automated Vehicles, highlighting their individual contributions to their fields and exploring the potential of their integration. Specifically, the objectives of this survey include mapping out the historical development of both technologies, examining current applications and integrations, and speculating on future directions and innovations at their intersection. By providing a comprehensive overview of the state of the art and identifying gaps in current research, this survey seeks to pave the way for future studies and technological breakthroughs in the confluence of AI and automotive technologies.

% We 
% \begin{itemize}
%     \item ...
%     \item ...
%     \item ...
% \end{itemize}

\begin{figure*}
  \centering
  \includegraphics*[width=1\textwidth]{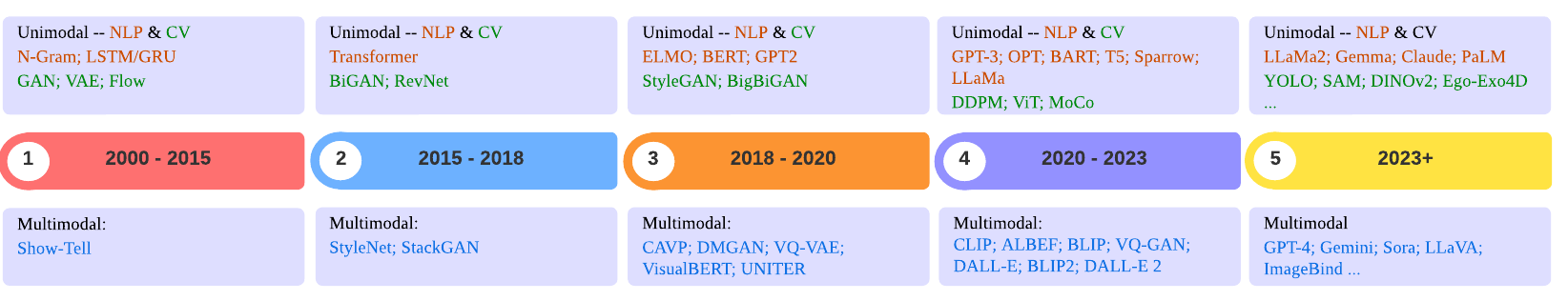}
  % \vspace{-25t}
  \caption{\textbf{The figure shows the development history of the generative model}}
  \label{model_history}
  \vspace{-10pt}
\end{figure*}

\section{Related Work}

\subsection{History of Generative Models}

The history of Generative Models and Connected and Automated Vehicles (CAVs) provides a rich context for understanding their potential intersection and future implications. Generative Models have evolved significantly over decades, from early innovations in procedural content generation \cite{b67} and Bayesian networks \cite{b60} to the development of deep learning techniques and architectures like Convolutional Neural Networks (CNNs) \cite{b61}, Recurrent Neural Networks (RNNs) \cite{b62}, and Generative Adversarial Networks (GANs) \cite{b63}. These models have found applications across various domains, including image and text generation, design, and simulation.

The development of Generative Models began with foundational work in AI and machine learning, including the LISP programming language in the 1960s, the ELIZA chatbot \cite{b64}, and early expert systems like Dendral \cite{b65} and MYCIN \cite{b66}. The rise of the internet and advancements in computing power in the 1990s and 2000s led to significant progress in machine learning, neural networks, and deep learning, setting the stage for modern generative AI \cite{b68}. Figure \ref{model_history} showcases a significant evolution from unimodal approaches in natural language processing (NLP) and computer vision (CV) towards increasingly sophisticated multimodal technologies. Early models like N-Gram \cite{b69} and GANs laid the groundwork between 2000 and 2015. The period from 2015 to 2018 saw the introduction of transformative architectures such as Transformers \cite{b70} and the emergence of multimodal models like StyleNet \cite{b71}. This evolution accelerated from 2018 to 2020 with advancements like BERT \cite{b72}, GPT-2 \cite{b73}, and StyleGAN, expanding to complex multimodal approaches including VisualBERT \cite{b74}. The trend from 2020 to 2023 highlights the proliferation of large language models like GPT-3 \cite{b75} and innovative visual technologies such as DALL-E \cite{b76}. From 2023 onward, we have witnessed a steady rise of innovative models like the remarkable GPT-4 \cite{b77}. However, the core architectures of LLMs have largely stabilized, and now research attention has increasingly shifted toward the development of Multimodal Large Language Models (MLLMs). This includes the rise of large vision-language models \cite{b98, b99, b100} capable of processing image-text pairs, followed by more general ``any-to-text'' models that accept inputs across modalities—such as images, video, audio, and sensor data—and generate coherent text outputs \cite{b101}. The latest frontier involves ``any-to-any'' generation models that can flexibly take in and produce outputs across multiple modalities \cite{b102, b103, b104}.

As both LLMs and MLLMs continue to mature, researchers are increasingly integrating them into agentic frameworks \cite{b105, b106}, enabling these models to interact with external tools and environments. In this paradigm, LLMs and MLLMs act as the ``brains'', capable of reasoning, planning, and language understanding, while the agentic system provides the ``arms,'' granting them the ability to perceive the world, execute actions, and call APIs or retrieve external knowledge. This combination marks a transformative shift from static model usage toward dynamic, goal-directed agents. For example, in the context of autonomous driving, these agents can interpret visual scenes, reason about traffic conditions, plan maneuvers, and take appropriate actions in real time. Models such as Sora \cite{b78}, DriveGPT \cite{b107}, and Genie \cite{b108} exemplify this integration by coupling generative understanding with action-oriented execution. This synergy between multimodal understanding and interactive agency is paving the way for next-generation intelligent systems that are both perceptive and capable of autonomous decision-making in complex real-world environments.

% period saw the introduction of pivotal technologies like GANs and transformers \cite{}, which have propelled forward the capabilities of generative AI, including the creation of sophisticated models like GPT \cite{} and DALL-E \cite{}.

\subsection{Challenges in the Generative Models}

Despite the advancements outlined in the previous section, generative models still confront a host of unresolved challenges that span ethical, legal, and technical domains. A prominent issue lies in the ethical considerations surrounding data privacy, biases in the training data, and the potential for misuse in creating deepfakes or spreading misinformation. The ethical dilemmas extend to copyright and legal exposure, as these models are trained on vast databases of images and text from various sources, raising concerns about intellectual property infringement and the legal repercussions of data use \cite{b79}.

Efforts have been made to mitigate the generation of inappropriate information through strategies like jailbreak and prompt injection \cite{b4}. However, malicious entities continue to devise new methods to exploit generative models, highlighting a persistent security threat\cite{b5, b6}. The rise in these attacks complicates the use of comprehensive datasets for training, as fears of revealing sensitive or harmful information loom large. 

A promising approach to addressing data privacy challenges involves developing more sophisticated algorithms to counteract malicious inputs. Research initiatives like Tensor Trust \cite{b18} have engaged in creating defenses against prompt injections through an interactive online game, generating a significant dataset with over 126,000 attacks and 46,000 defenses. Additionally, Jatmo \cite{b19} has introduced a novel method for constructing task-specific models that are inherently resistant to prompt injections by leveraging a teacher model for generating tailored datasets. This advancement demonstrates a critical step forward in enhancing generative models' ability to autonomously identify and mitigate harmful inputs, thus bolstering data privacy protections.

Furthermore, the phenomenon of model hallucination, where generative models fabricate information not present in their training data, underscores the challenge of ensuring reliability \cite{b80}. While approaches like Retrieval-Augmented Generation (RAG) \cite{b81} and fine-tuning \cite{b82} offer some solutions, they introduce additional complexities such as increased time and computational costs. 

One way to improve the computational cost of fine-tuning is by utilizing Low Rank Adaptor (LoRA) \cite{b83}, which introduces trainable parameters that capture important information in a lower-dimensional space. This method modifies only a small portion of the model's weights, reducing the number of parameters that need to be updated during fine-tuning. By focusing on these adaptable components, LoRA efficiently updates the model, maintaining performance while significantly lowering computational demands and memory usage.

Improving the performance of Retrieval Augmented Generation (RAG) involves several strategic enhancements across data preparation, indexing, and query handling. To reduce computational time, we can explore various index types for better context retrieval. Additionally, we can also transform queries to better match the retrieval context. Each of these tactics aims at refining the interaction between the LLM and the data, ensuring more accurate, relevant, and efficient generation outcomes \cite{b20}.

% Moreover, the training of generative AI on the creative outputs of millions of individuals without compensation poses significant ethical and legal questions. The capability of these models to produce highly realistic and persuasive content can be weaponized for misinformation campaigns, posing a threat to democratic discourse and public trust. The socio-ethical implications of generative AI demand a careful examination of its impacts on human knowledge, creativity, and the equitable distribution of its benefits.

% Despite the advancements from the previous section, generative models face a slew of unsolved challenges. One of the major issues is the ethical considerations surrounding data privacy, biases in training data, and the potential for misuse in creating deepfakes or propagating misinformation \cite{}. Even though researchers have create strategies to prevent models to generate inappropriate information, such as jailbreak, prompt injection \cite{b4}, malicious people are still coming up with new way to attack generative models \cite{b5, b6}. With the rise of being attack, people can no longer provide all the information when training the model, since they are afraid of sensitive information reveal. Additionally, the model hallucination is also a huge challenge that cause issue of reliability. Even though there are some solution to this challenge, such as RAG \cite{}, finetune \cite{}, etc. It still make the model to suffer from time complexity and the cost.

\begin{table*}[ht]
\centering % centers the table
\caption{Advantage and Disadvantage of Generative Modle in AV}
\label{tab:sample_table_lines}

\noindent % Ensures the tabular starts at the left margin
\begin{tabular}{| m{3cm} | m{7cm} | m{6.8cm} |}
    \hline
    \textbf{Model} & \textbf{Advantages} & \textbf{Disadvantages} \\ \hline
    Generative Adversarial Networks (GAN) \cite{b7} \cite{b8}& \textbf{Realistic Data Generation:} GANs can produce highly realistic synthetic data, aiding in diverse scenario training for automotive vehicles (AV) systems without costly real-world data collection. \newline \textbf{Data Augmentation:} Enable the enhancement of existing datasets with varied conditions, crucial for comprehensive AV system training. \newline \textbf{Anomaly Detection:} Capable of identifying anomalies by learning normal operational patterns, enhancing safety mechanisms in AVs. & 
    
    \textbf{Training Complexity:} GANs are challenging to train, often facing issues like mode collapse, where the diversity of generated samples is limited.
    \newline
    \textbf{High Computational Demand:}The generation of high-quality data through GANs requires substantial computational resources.
    Bias Propagation: Biases in training data can be mirrored in the generated data, possibly leading to biased learning outcomes in AVs.
    
    \\ 
    \hline
    Reinforcement Learning \cite{b9} \cite{b10}& 
    \textbf{Adaptive Decision Making:} RL models are excellent at learning optimal actions through trial and error, enabling autonomous vehicles to adapt to changing road conditions dynamically.\newline
    \textbf{Continuous Learning:} Continuously improve by learning from interactions with the environment, enhancing the performance and safety of autonomous vehicles over time.
     &
    \textbf{Sample Efficiency:} RL models often require a significant number of interactions with the environment, making the learning process resource-intensive and time-consuming.\newline
    \textbf{Complexity and Scalability:} Designing RL algorithms that perform well across various driving scenarios is challenging, which can limit the scalability and general applicability of these models in complex environments.
    \\ 
    \hline
    StyleGAN \cite{b11}\cite{b12}& 
    \textbf{High-Quality Images:} Produces high-resolution, photo realistic images with fine details.\newline
    \textbf{Control Over Generation:} Offers control over specific features of the generated images through style-based generation, allowing for detailed customization.\newline
    \textbf{Variety and Diversity:} Capable of generating a wide variety of images within the same framework, showcasing impressive diversity.\newline
     &
   \textbf{Complexity and Resources:} Requires significant computational resources and expertise to train, limiting accessibility.\newline
    \textbf{Training Difficulties:} Can encounter stability issues during training, requiring careful tuning of parameters.\newline
   \textbf{ Potential for Misuse:} High-quality synthetic image generation raises ethical concerns, including the creation of deepfakes.

    \\ \hline
    Neural Architecture Search(NAS) \cite{b13}\cite{b14}& 

    \textbf{Automation: }NAS automates the design of network architectures, potentially outperforming manually designed networks, especially in multi-objective optimization scenarios.\newline
    \textbf{Efficiency:} It enables the discovery of novel network architectures optimized for specific hardware constraints, improving sensor fusion performance and efficiency on embedded devices.
     &
   \textbf{Time Consuming: }NAS processes can be computationally intensive and time-consuming, requiring significant resources for training and evaluation of numerous architectural configurations.\newline
    \textbf{Complexity Balance:} There might be a complexity in balancing the trade-offs between model size, performance, and computational efficiency, especially under strict hardware constraints.
    
   \\ \hline
    Collaborative AI \cite{b15}\cite{b16}\cite{b17}& 
    \textbf{Enhanced Learning and Adaptation:} Collaborative AI allows vehicles to learn from each other's experiences, significantly improving their ability to adapt to new environments and situations without direct human intervention.\newline
    \textbf{Increased Data Diversity:} It facilitates access to a broader range of data collected from various vehicles operating in different conditions, leading to more robust and generalizable AI models.\newline
    \textbf{Efficiency in Data Use:} By sharing insights rather than raw data, collaborative AI can efficiently utilize bandwidth and storage, ensuring timely updates and learning without overwhelming the system's resources.
    \newline
    \textbf{Improved Safety and Reliability:}Vehicles can benefit from shared knowledge about hazardous conditions, traffic congestion, and road safety, leading to more informed decision-making and enhanced safety for all road users.
     &
    \textbf{Data Privacy and Security:} Collaborating and sharing data between vehicles raise concerns about user privacy and data security. Ensuring the integrity and confidentiality of shared information is critical.
    \newline
    \textbf{System Complexity and Integration:} Implementing collaborative AI requires sophisticated systems capable of managing communication, data processing, and learning across different vehicles and infrastructure, adding complexity to the autonomous driving ecosystem.
    \newline
    \textbf{Dependency on Connectivity:} The effectiveness of collaborative AI hinges on reliable connectivity. Issues such as signal loss, latency, or network failures could impact the system's performance and safety.
    \newline
    \textbf{Standardization and Compatibility:} Achieving seamless collaboration requires standardized protocols and interfaces across different manufacturers and models. Lack of standardization can limit interoperability and the overall effectiveness of collaborative AI systems.

   \\ \hline
       
\end{tabular}

\end{table*}

\begin{figure*}
  \centering
  \includegraphics*[width=1\textwidth]{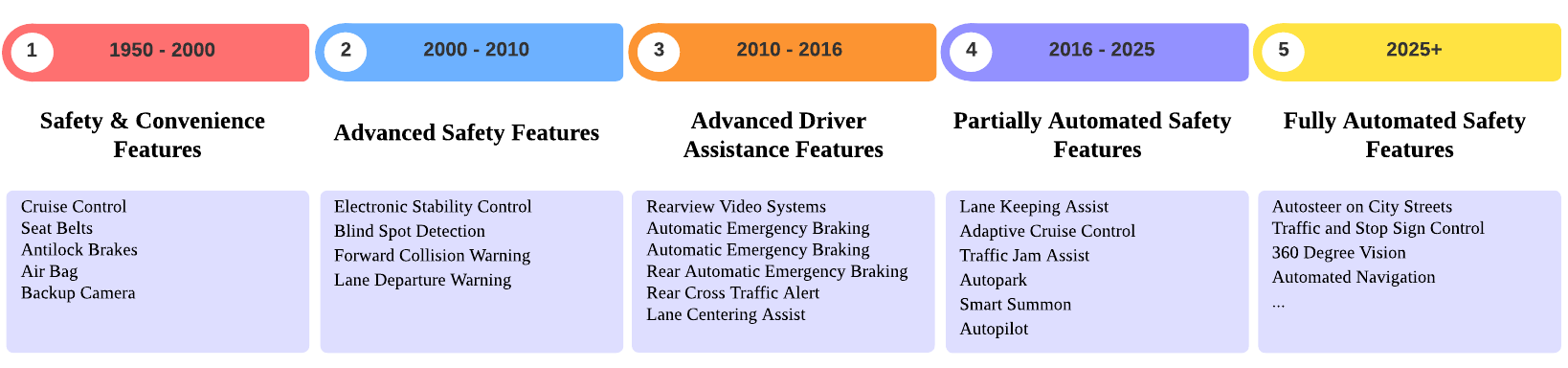}
  % \vspace{-25t}
  \caption{\textbf{The figure shows the development history of the car safety}}
  \label{challenge_history}
  \vspace{-10pt}
\end{figure*}

\subsection{History of Connected and Automated Vehicles (CAVs)}

The concept of connected cars has been around since the mid-1990s, with General Motors' introduction of OnStar in 1996 marking a significant early milestone \cite{b84}. This system, developed in collaboration with Motorola Automotive, aimed primarily at enhancing vehicle safety and providing emergency services. Since then, the scope of connected car features has expanded significantly to include mobility management, commerce, vehicle management, safety, entertainment, driver assistance, well-being, and breakdown prevention. Innovations such as Google's formation of the Open Automotive Alliance in 2014 \cite{b85} and the launch of Apple's CarPlay \cite{b86} and Android Auto \cite{b87} signify the growing integration of smartphone technology with vehicle infotainment systems. This evolution underscores a shift towards enhancing driver experience, safety, and vehicle efficiency through connectivity.

On the other hand, the development of autonomous vehicles (AVs) represents a parallel trajectory towards reducing the need for human intervention in vehicle operation \cite{b88}. The Society of Automotive Engineers (SAE) defines six levels of automation for vehicles, ranging from no automation (Level 0) to full automation (Level 5), where the vehicle is capable of performing all driving functions under all conditions without human input. The current state of technology primarily falls between Levels 3 and 4, where vehicles can perform some driving functions independently but still require human oversight. The technology underpinning AVs includes radar, GPS, cameras, and lidar to create a detailed 3D map of the vehicle's surroundings, enabling decision-making and vehicle control through advanced computer systems, machine learning, and artificial intelligence \cite{b88}.

As of recent developments, the industry continues to face challenges, including regulatory hurdles, technological limitations, and public skepticism. Incidents involving self-driving car companies like Waymo highlight the ongoing issues related to safety and public acceptance of autonomous technology \cite{b89}. However, efforts such as dedicated lanes for CAVs and advancements in vehicle-to-vehicle (V2V) \cite{b90} and vehicle-to-infrastructure (V2I) \cite{b91} communications demonstrate a clear commitment to overcoming these obstacles and pushing the boundaries of what's possible in smart transportation.

\subsection{Challenges in the Connected and Automated Vehicles}

The journey towards fully autonomous vehicles is fraught with challenges, chief among them being safety and reliability. While the promise of accident-free mobility and significant reductions in road fatalities is the motivation behind CAVs, we realize this goal is too complex \cite{b92}. The National Highway Traffic Safety Administration (NHTSA) outlines the stages of automation from Level 0 (no automation) to Level 5 (full automation), with current consumer technologies mainly falling between Levels 2 and 3. These levels highlight the incremental steps towards fully autonomous systems, where the vehicle is responsible for all driving tasks within certain conditions (Level 3) to all conditions (Level 5)\cite{b93}. However, these advanced driving systems, crucial for removing the human driver from the chain of events leading to a crash, are not yet available for consumer purchase, underscoring the gap between current capabilities and the goal of full automation \cite{b94}.

Looking back at the history of vehicle safety, we've seen tremendous progress through various challenges on the path towards fully autonomous driving. This journey can be mapped through the "Five Eras of Safety" as outlined by the NHTSA \cite{b1}. As figure \ref{challenge_history} shows, these eras highlight the evolution from basic manual safety features to the sophisticated, automated systems that are paving the way for fully autonomous vehicles. Each era has brought with it significant advancements in technology and regulation, from the introduction of seat belts and airbags to the development of Safety and Convenience Features, to the brink of Fully Automated Safety Features. This historical perspective underscores the collaborative efforts between automakers, technology companies, and regulatory bodies in overcoming obstacles and innovating towards a safer automotive future.

% Despite the previous mentioned success, we are still facing lots of unsolved challenge. The furture we go, the more difficult the challenge become. 
Despite the significant progress and overcoming of numerous challenges on the way to fully autonomous driving, we are currently facing a new set of challenges that appear to grow more complex as we advance further. One of the major challenges includes Autosteer on City Streets, where vehicles must navigate complex urban environments, recognizing and responding to traffic signs, signals, and unpredictable human behaviors. This complexity is compounded by the requirement for Traffic and Stop Sign Control, where vehicles must accurately identify and react to stop signs and traffic lights in real-time, ensuring safe and law-compliant driving \cite{b95}. Moreover, achieving 360 Degree Vision is pivotal for autonomous vehicles to ensure a comprehensive understanding of their surroundings, enabling them to detect obstacles, pedestrians, and other vehicles from every angle. This is essential for safe navigation, especially in densely populated urban areas. However, developing such sophisticated sensor systems that can reliably function under various weather and lighting conditions presents significant technical and financial challenges \cite{b96}. Automated Navigation poses another significant challenge, requiring advanced algorithms capable of planning optimal routes in real-time while considering current traffic conditions, road works, and other dynamic factors \cite{b97}.

The challenges extend beyond technical capabilities, touching on infrastructure and regulatory frameworks. The infrastructure needs to evolve to support autonomous vehicles fully, requiring clear lane markings, reliable Vehicle-to-Infrastructure (V2I) communication systems, and robust data storage solutions \cite{b2}. Regulatory support is crucial to address safety concerns, establish trusted ecosystems, and implement global standards. This includes updates to road maintenance practices and the introduction of new funding models to support the necessary infrastructure upgrades without significantly impacting public budgets \cite{b3}.

% TODO
\section{Integration of Generative Models in CAVs}
% TODO 介绍实际应用 
% VistaGPT: Generative Parallel Transformers for Vehicles with Intelligent Systems for Transport Automation
% A Systematic Solution of Human Driving Behavior Modeling and Simulation for Automated Vehicle Studies
% On the Integration of Enabling Wireless Technologies and Sensor Fusion for Next-Generation Connected and Autonomous Vehicles
% Safe Model-Based Off-Policy Reinforcement Learning for Eco-Driving in Connected and Automated Hybrid Electric Vehicles
\subsection{Integration in Real Life}

In the field of Connected Automated Vehicles (CAV), as Table \ref{tab:sample_table_lines} shows various computational models like Generative Adversarial Networks (GANs), Reinforcement Learning (RL), StyleGAN, Neural Architecture Search (NAS), and Collaborative AI significantly enhance AV intelligence and safety. GANs contribute by generating synthetic data for diverse scenario training, though they are complex to train and may propagate bias \cite{b25}. Creswell et al. \cite{b29} excels in adaptive decision-making but is resource-intensive in RL. Shalev-Shwartz et al. \cite{b31} offers high-resolution image generation for training data but requires substantial resources and poses ethical risks in StyleGAN. Karras et al.\cite{b34} streamlines network architecture design, optimizing for specific constraints yet demanding in terms of computational resources in NAS. Tan et al. \cite{b37} facilitates shared learning and data diversity among vehicles, improving adaptability and model robustness, albeit raising concerns over data privacy and the need for reliable connectivity in Collaborative AI. Despite these challenges, such as computational demands and ethical considerations, the benefits of these models in improving safety, efficiency, and adaptability are undeniable, underscoring the need for ongoing advancements to fully leverage their potential in CAV technology  \cite{b40}. Here are some real life application examples.

\subsubsection{VistaGPT}
VistaGPT \cite{b46} leverages the capabilities of generative models to enhance traffic management, particularly at congested urban intersections. By analyzing extensive traffic data, including vehicle speeds and pedestrian movements, VistaGPT predicts traffic patterns, enabling dynamic optimization of traffic light timings. This reduces congestion and wait times, showcasing the potential of AI in improving urban mobility and efficiency.

The practical efficacy of VistaGPT was rigorously tested through a pilot project undertaken in a densely populated metropolitan area, where the system was seamlessly incorporated into the existing traffic management infrastructure. The outcomes of this integration were profound, with the project documenting a substantial reduction in wait times at critical intersections by up to 25\% during peak traffic periods. This improvement in traffic flow not only underscored VistaGPT's capability to significantly enhance urban traffic management but also highlighted its environmental impact through the reduction of vehicular emissions attributed to prolonged idling at traffic stops. Moreover, VistaGPT's predictive functionality ensures that the traffic management system can respond proactively to unexpected traffic conditions, such as accidents or emergency vehicle prioritization, further underscoring the system's value in creating more adaptable and responsive urban transportation networks. The successful deployment of VistaGPT in this real-world scenario signals a promising direction for the future of intelligent transportation systems, where AI-driven solutions can lead to safer, more efficient, and environmentally friendly urban environments.

\subsubsection{Solution of Human Driving Behavior Modeling}
The integration of systematic human driving behavior modeling and simulation into automated vehicle (AV) studies presents a groundbreaking approach to enhancing the interaction between human drivers and autonomous systems. A pivotal application of this methodology is observed in the development of a virtual simulation environment designed to mirror the complexities of real-world driving scenarios. This environment employs advanced behavioral models to accurately represent a wide array of human driving behaviors, such as aggressive and cautious driving patterns, as well as unpredictable human actions on the road. The primary aim of this initiative was to assess and refine the adaptability and responsiveness of AVs when navigating mixed-traffic environments, which are characterized by the coexistence of human-operated vehicles and AVs \cite{b47}.

The project yielded remarkable insights, particularly in the domain of improving safety protocols and traffic efficiency for AVs operating alongside human drivers. By simulating diverse human driving behaviors and their potential impact on road safety, researchers were able to enhance the decision-making algorithms of AVs, enabling these vehicles to anticipate human actions with greater precision and modify their operation to avert accidents. The findings from this study revealed that AVs equipped with these enhanced algorithms could significantly diminish the likelihood of traffic incidents, with simulations showing up to a 30\% reduction in accidents in mixed-traffic conditions. This underscores the vital role that understanding human driving behavior plays in the evolution of autonomous driving technologies, emphasizing the effectiveness of simulation-based strategies in fostering the safe cohabitation of AVs and human drivers on public roads.

\subsubsection{Integrating Wireless Technologies and Sensor Fusion in CAVs}
The integration of enabling wireless technologies and sensor fusion is transforming the landscape of next-generation Connected and Autonomous Vehicles (CAVs), with practical applications already emerging in smart city infrastructures. A notable project in this realm focused on leveraging Dedicated Short-Range Communications (DSRC) and the burgeoning 5G networks to facilitate advanced Vehicle-to-Everything (V2X) communications. This synergy, coupled with sensor fusion that harmonizes data inputs from LiDAR, radar, and cameras, equips CAVs with unparalleled situational awareness. For example, in a pilot implementation in a metropolitan area, this integration enabled CAVs to navigate complex urban terrains by detecting obstacles, traffic, and pedestrian movements in real-time, significantly enhancing safety and traffic efficiency \cite{b48, b49}.

Further, this technological amalgamation has pioneered new paradigms in traffic management and vehicle coordination. In scenarios such as intersection crossing, CAVs utilize these wireless and sensor fusion technologies to communicate with each other and with traffic infrastructure to optimize traffic flow and reduce wait times, effectively minimizing the reliance on traditional traffic control devices. This application not only illustrates the potential of these technologies to streamline urban transportation but also highlights their role in mitigating traffic congestion and fostering a sustainable urban mobility ecosystem. The advancements documented in projects like these underscore the critical importance of continued innovation in wireless communication and sensor technologies for the evolution of autonomous driving and the realization of fully connected and intelligent transportation systems \cite{b50, b51}.

\subsubsection{Eco-Driving through AI in Hybrid Electric Vehicles}
The deployment of Safe Model-Based Off-Policy Reinforcement Learning for enhancing eco-driving in Connected and Automated Hybrid Electric Vehicles (CAV-HEVs) has made notable strides in improving fuel efficiency and reducing environmental impact. In a key project, researchers developed a model leveraging off-policy reinforcement learning to optimize driving behaviors and powertrain operations for fuel savings, utilizing real-time data from V2V and V2I communications. This model enabled CAV-HEVs to dynamically adjust to live traffic and environmental conditions, promoting efficient route selection and vehicle operation.

A field trial involving a fleet of CAV-HEVs showcased a substantial 20\% reduction in fuel consumption compared to traditional driving methods, while maintaining high safety standards. This achievement highlights the potential of integrating advanced AI algorithms with eco-driving techniques to promote sustainable automotive technologies. The project exemplifies how intelligent vehicle systems can contribute to environmental sustainability goals by optimizing energy usage in urban transportation \cite{b52, b53}.

\subsection{Future Directions}

\subsubsection{Perception and scene understanding}
Future directions for integrating generative models with Connected and Automated Vehicles (CAVs) are poised to significantly enhance perception and scene understanding capabilities, a foundational aspect for the advancement of autonomous driving technologies. As vehicles evolve to interpret their environments with greater accuracy, real-time recognition and response to both static and dynamic elements become imperative. While the work by Muhammad et al. (2022) \cite{b21} explores advancements in vision-based technologies for autonomous driving, it also highlights significant challenges that impede optimal performance. Notably, existing limitations, such as the oversight of locational context during classification, diminished performance under adverse weather conditions, and the underutilization of vision transformers, underscore the necessity for continued innovation in this field. Addressing these challenges will not only refine the current approaches but also unlock new potentials for generative models to revolutionize how CAVs perceive and interact with their surroundings, marking a significant leap forward in the quest for fully autonomous driving systems.

\subsubsection{Prediction of other road users' behavior}
Beyond achieving comprehensive awareness and understanding of their surroundings, the future of CAVs also hinges on the ability to anticipate the actions of other road users. This predictive capability is crucial for ensuring smooth and safe interactions on the road, especially in complex scenarios such as urban intersections. For instance, when a vehicle signals a lane change through its left turn light, CAVs should be able to infer that the vehicle is likely to merge into their lane and adjust their behavior accordingly. Kalatian et al. \cite{b22} sheds light on significant advancements in CAV technologies. This study puts forward a context-aware model utilizing virtual reality data to simulate pedestrian behavior, particularly at mid-block unsignalized crossings. By integrating a multi-input network of Long Short-Term Memory (LSTM) and fully connected dense layers, the model incorporates not just past trajectories but also pedestrian head orientations and their distance to approaching vehicles as sequential input data. The study also acknowledges the limitations of this approach, including challenges in accurately capturing the dynamic interactions between pedestrians and vehicles in various environmental conditions and the need for extensive data to train the models effectively. One of the future approach is to improve model accuracy under diverse scenarios, such as different weather conditions, varied pedestrian behaviors, and complex urban landscapes.

\subsubsection{Enhanced Decision-Making}
% Other than understanding and prediction, vehicle and models should also able to make decision of what to do next based on the prediction. Also, this decision should be the optimal and safest decision ever. Hang et al. \cite{b23} proposed a game-theoretic decision-making framework aimed at enhancing the coordination of connected automated vehicles (CAVs) at urban intersections. This framework is designed to maximize both the social benefits, such as traffic system efficiency and safety, and the benefits of individual users. At the core of this decision-making process is the unsignalized intersection scenario, where vehicles must collaboratively decide on their actions without relying on traffic signals. The decision-making algorithm integrates a Gaussian potential field approach to assess driving risks, helping to simplify the complexity of real-time decision-making. the study acknowledges limitations, including potential challenges in addressing the dynamic interactions between vehicles and the environment, the necessity for extensive data to train the models, and the optimization of the algorithm to ensure efficiency and safety in various driving conditions.

Beyond mere perception and predictive capabilities, vehicles and their corresponding models must also excel in making decisions about subsequent actions based on these predictions. Such decisions should represent the pinnacle of safety and optimality. Hang et al. \cite{b23} introduced a game-theoretic framework specifically designed to improve the coordination of Connected Automated Vehicles (CAVs) at urban intersections, targeting the augmentation of both communal benefits, like traffic system efficiency and safety, and individual user advantages. Central to this framework is the challenge presented by unsignalized intersections, where vehicles are required to collaboratively make decisions without traffic signal guidance. Incorporating a Gaussian potential field approach for risk assessment, this framework aims to reduce the complexity inherent in real-time decision-making. In the future, researchers should continue on this path to solve the limitations that Hang et all. proposed, such as difficulties in fully capturing dynamic vehicle-environment interactions, the extensive dataset necessary for model training, and the need to refine the algorithm for enhanced efficiency and safety across varied driving scenarios.

\section{Conclusion}

This survey has looked into combining Generative Models with Connected and Automated Vehicles (CAVs). It has shown progress and obstacles in artificial intelligence and autonomous transportation. Our study found positive connections between generative models and CAVs, such as improving predictive modeling, simulation accuracy, and decision-making for autonomous vehicles.

Throughout the survey, we identified critical advancements in generative models, such as Generative Adversarial Networks (GANs), Reinforcement Learning, StyleGAN, Neural Architecture Search (NAS), and Collaborative AI, each offering unique contributions to enhancing the intelligence, safety, and efficiency of CAVs. Despite these advancements, the integration of generative models into CAVs faces challenges, including ethical considerations, data privacy concerns, computational demands, and the reliability of generated data.

Real-world applications, such as VistaGPT for traffic management, systematic human driving behavior modeling, the integration of wireless technologies and sensor fusion in CAVs, and AI-driven eco-driving in hybrid electric vehicles, demonstrate the practical benefits and potential of leveraging generative models in the context of CAVs. These applications not only improve safety and efficiency but also pave the way for innovative solutions in smart transportation systems.

Looking ahead, the future of CAVs will depend on overcoming the current challenges and further harnessing the power of generative models. This includes enhancing perception and scene understanding, improving the prediction of other road users' behavior, and advancing decision-making algorithms for autonomous vehicles. Addressing these areas will require a multidisciplinary approach, combining expertise from artificial intelligence, automotive engineering, ethics, and policy-making, to fully realize the potential of CAVs and ensure their safe, efficient, and ethical integration into our transportation systems.

In conclusion, the integration of Generative Models with CAVs holds tremendous potential for revolutionizing the transportation industry. By continuing to address the challenges and harness the opportunities presented by this synergy, we can look forward to a future where autonomous vehicles operate more safely, efficiently, and harmoniously within our transportation ecosystems.

% \section*{Acknowledgment}

% \section*{References}

% Please number citations consecutively within brackets \cite{b1}. The 
% sentence punctuation follows the bracket \cite{b2}. Refer simply to the reference 
% number, as in \cite{b3}---do not use ``Ref. \cite{b3}'' or ``reference \cite{b3}'' except at 
% the beginning of a sentence: ``Reference \cite{b3} was the first $\ldots$''

% Number footnotes separately in superscripts. Place the actual footnote at 
% the bottom of the column in which it was cited. Do not put footnotes in the 
% abstract or reference list. Use letters for table footnotes.

% Unless there are six authors or more give all authors' names; do not use 
% ``et al.''. Papers that have not been published, even if they have been 
% submitted for publication, should be cited as ``unpublished'' \cite{b4}. Papers 
% that have been accepted for publication should be cited as ``in press'' \cite{b5}. 
% Capitalize only the first word in a paper title, except for proper nouns and 
% element symbols.

% For papers published in translation journals, please give the English 
% citation first, followed by the original foreign-language citation \cite{b6}.

% \vspace{12pt}
% \color{red}
% IEEE conference templates contain guidance text for composing and formatting conference papers. Please ensure that all template text is removed from your conference paper prior to submission to the conference. Failure to remove the template text from your paper may result in your paper not being published.


\begin{thebibliography}{00}
\bibitem{b1} NHTSA, “Automated Vehicles for Safety | NHTSA,” www.nhtsa.gov. https://www.nhtsa.gov/vehicle-safety/automated-vehicles-safety
\bibitem{b2} McKinsey, “Autonomous driving’s future: Convenient and connected | McKinsey,” www.mckinsey.com, Jan. 06, 2023. https://www.mckinsey.com/industries/automotive-and-assembly/our-insights/autonomous-drivings-future-convenient-and-connected
‌\bibitem{b3} R. McCauley, “The 6 Challenges of Autonomous Vehicles and How to Overcome Them,” Govtech.com, 2019. https://www.govtech.com/fs/The-6-Challenges-of-Autonomous-Vehicles-and-How-to-Overcome-Them.html
‌‌\bibitem{b4} Zhou, Andy, Bo Li, and Haohan Wang. "Robust prompt optimization for defending language models against jailbreaking attacks." arXiv preprint arXiv:2401.17263 (2024).
\bibitem{b5} Yu, Jiahao, et al. "Assessing prompt injection risks in 200+ custom gpts." arXiv preprint arXiv:2311.11538 (2023).
\bibitem{b6} Jin, Mingyu, et al. "AttackEval: How to Evaluate the Effectiveness of Jailbreak Attacking on Large Language Models." arXiv preprint arXiv:2401.09002 (2024).
\bibitem{b7} Divya Saxena and Jiannong Cao. 2021. Generative Adversarial Networks (GANs): Challenges, Solutions, and Future Directions. ACM Comput. Surv. 54, 3, Article 63 (April 2022), 42 pages. https://doi.org/10.1145/3446374
\bibitem{b8} H. Lin, Y. Liu, S. Li and X. Qu, "How Generative Adversarial Networks Promote the Development of Intelligent Transportation Systems: A Survey," in IEEE/CAA Journal of Automatica Sinica, vol. 10, no. 9, pp. 1781-1796, September 2023, doi: 10.1109/JAS.2023.123744.
\bibitem{b9}S. Aradi, "Survey of Deep Reinforcement Learning for Motion Planning of Autonomous Vehicles," in IEEE Transactions on Intelligent Transportation Systems, vol. 23, no. 2, pp. 740-759, Feb. 2022, doi: 10.1109/TITS.2020.3024655
\bibitem{b10}A. Alharin, T. -N. Doan and M. Sartipi, "Reinforcement Learning Interpretation Methods: A Survey," in IEEE Access, vol. 8, pp. 171058-171077, 2020, doi: 10.1109/ACCESS.2020.3023394.
\bibitem{b11}Rishubh Parihar, Ankit Dhiman, Tejan Karmali, and Venkatesh R. 2022. Everything is There in Latent Space: Attribute Editing and Attribute Style Manipulation by StyleGAN Latent Space Exploration. In Proceedings of the 30th ACM International Conference on Multimedia (MM '22). Association for Computing Machinery, New York, NY, USA, 1828–1836. https://doi.org/10.1145/3503161.3547972
\bibitem{b12}Omer Tov, Yuval Alaluf, Yotam Nitzan, Or Patashnik, and Daniel Cohen-Or. 2021. Designing an encoder for StyleGAN image manipulation. ACM Trans. Graph. 40, 4, Article 133 (August 2021), 14 pages. https://doi.org/10.1145/3450626.3459838
\bibitem{b13}C. Hao et al., "NAIS: Neural Architecture and Implementation Search and its Applications in Autonomous Driving," 2019 IEEE/ACM International Conference on Computer-Aided Design (ICCAD), Westminster, CO, USA, 2019, pp. 1-8, doi: 10.1109/ICCAD45719.2019.8942055. 
\bibitem{b14}G. Balazs and W. Stechele, "Neural Architecture Search for Automotive Grid Fusion Networks Under Embedded Hardware Constraints," 2020 19th IEEE International Conference on Machine Learning and Applications (ICMLA), Miami, FL, USA, 2020, pp. 79-86, doi: 10.1109/ICMLA51294.2020.00022.
\bibitem{b15}T. Zeng, O. Semiari, M. Chen, W. Saad and M. Bennis, "Federated Learning for Collaborative Controller Design of Connected and Autonomous Vehicles," 2021 60th IEEE Conference on Decision and Control (CDC), Austin, TX, USA, 2021, pp. 5033-5038, doi: 10.1109/CDC45484.2021.9683257. 
\bibitem{b16}Z. Xiao, J. Shu, H. Jiang, G. Min, J. Liang and A. Iyengar, "Toward Collaborative Occlusion-free Perception in Connected Autonomous Vehicles," in IEEE Transactions on Mobile Computing, doi: 10.1109/TMC.2023.3298643.
\bibitem{b17}Julio C. S. Dos Anjos, Kassiano J. Matteussi, Fernanda C. Orlandi, Jorge L. V. Barbosa, Jorge Sá Silva, Luiz F. Bittencourt, and Cláudio F. R. Geyer. 2023. A Survey on Collaborative Learning for Intelligent Autonomous Systems. ACM Comput. Surv. 56, 4, Article 98 (April 2024), 37 pages. https://doi.org/10.1145/3625544
\bibitem{b18}Toyer, Sam, et al. "Tensor trust: Interpretable prompt injection attacks from an online game." arXiv preprint arXiv:2311.01011 (2023).
\bibitem{b19}Piet, Julien, et al. "Jatmo: Prompt injection defense by task-specific finetuning." arXiv preprint arXiv:2312.17673 (2023).
\bibitem{b20}M. Ambrogi, “10 Ways to Improve the Performance of Retrieval Augmented Generation Systems,” Medium, Sep. 18, 2023. https://towardsdatascience.com/10-ways-to-improve-the-performance-of-retrieval-augmented-generation-systems-5fa2cee7cd5c
\bibitem{b21}Muhammad, Khan, et al. "Vision-based semantic segmentation in scene understanding for autonomous driving: Recent achievements, challenges, and outlooks." IEEE Transactions on Intelligent Transportation Systems 23.12 (2022): 22694-22715.
\bibitem{b22}Kalatian, Arash, and Bilal Farooq. "A context-aware pedestrian trajectory prediction framework for automated vehicles." Transportation research part C: emerging technologies 134 (2022): 103453.
\bibitem{b23}Hang, Peng, et al. "Decision making for connected automated vehicles at urban intersections considering social and individual benefits." IEEE transactions on intelligent transportation systems 23.11 (2022): 22549-22562.
\bibitem{b24}Goodfellow, I. J., Pouget-Abadie, J., Mirza, M., Xu, B., Warde-Farley, D., Ozair, S.,and Bengio, Y. (2014). Generative adversarial nets. In Advances in neural information processing systems (pp. 2672-2680).
\bibitem{b25}Antoniou, A., Storkey, A., and Edwards, H. (2017). Data augmentation generative adversarial networks. arXiv preprint arXiv:1711.04340.
\bibitem{b26}Schlegl, T., Seeböck, P., Waldstein, S. M., Schmidt-Erfurth, U., and Langs, G. (2017). Unsupervised anomaly detection with generative adversarial networks to guide marker discovery. In International conference on information processing in medical imaging (pp. 146-157). Springer, Cham.
\bibitem{b27}Salimans, T., Goodfellow, I., Zaremba, W., Cheung, V., Radford, A., Chen, X., and Chen, Y. (2016). Improved techniques for training GANs. In Advances in neural information processing systems (pp. 2234-2242).
\bibitem{b28}Brock, A., Donahue, J., and Simonyan, K. (2019). Large scale GAN training for high fidelity natural image synthesis. In Proceedings of the International Conference on Learning Representations (ICLR).
\bibitem{b29}Creswell, A., White, T., Dumoulin, V., Arulkumaran, K., Sengupta, B., and Bharath, A. A. (2018). Generative adversarial networks: An overview. IEEE Signal Processing Magazine, 35(1), 53-65.
\bibitem{b30}Mnih, V., Kavukcuoglu, K., Silver, D., Rusu, A. A., Veness, J., Bellemare, M. G., and Petersen, S. (2015). Human-level control through deep reinforcement learning. Nature, 518(7540), 529-533.
\bibitem{b31}Shalev-Shwartz, S., Shammah, S., and Shashua, A. (2016). Safe, multi-agent, reinforcement learning for autonomous driving. arXiv preprint arXiv:1610.03295.
\bibitem{b32}Henderson, P., Islam, R., Bachman, P., Pineau, J., Precup, D.,and Meger, D. (2018). Deep reinforcement learning that matters. In Proceedings of the AAAI Conference on Artificial Intelligence (Vol. 32, No. 1).
\bibitem{b33}Dulac-Arnold, G., Mankowitz, D., and Hester, T. (2019). Challenges of real-world reinforcement learning. arXiv preprint arXiv:1904.12901.
\bibitem{b34}Karras, T., Laine, S.,and Aila, T. (2019). A style-based generator architecture for generative adversarial networks. In Proceedings of the IEEE/CVF Conference on Computer Vision and Pattern Recognition (pp. 4401-4410).
\bibitem{b35}Chesney, R., and Citron, D. K. (2019). Deep fakes: A looming challenge for privacy, democracy, and national security. California Law Review, 107, 1753.
\bibitem{b36}Zoph, B., and Le, Q. V. (2017). Neural architecture search with reinforcement learning. In Proceedings of the International Conference on Learning Representations (ICLR).
\bibitem{b37}Tan, M., Chen, B., Pang, R., Vasudevan, V., and Le, Q. V. (2019). Mnasnet: Platform-aware neural architecture search for mobile. In Proceedings of the IEEE/CVF Conference on Computer Vision and Pattern Recognition (pp. 2820-2828).
\bibitem{b38}Elsken, T., Metzen, J. H., and Hutter, F. (2019). Neural architecture search: A survey. Journal of Machine Learning Research, 20(55), 1-21.
\bibitem{b39}Shlezinger, N., Eldar, Y. C., and Fuhrmann, D. R. (2020). Model-based deep learning. arXiv preprint arXiv:2008.08414.
\bibitem{b40}Ran, C., Hu, X., Chen, Z., Sun, L., and Shi, J. (2019). Deep learning models for global collaborative autonomous driving. arXiv preprint arXiv:1909.05481.
\bibitem{b41}Sicari, S., Rizzardi, A., Grieco, L. A., and Coen-Porisini, A. (2015). Security, privacy and trust in Internet of Things: The road ahead. Computer networks, 76, 146-164.
\bibitem{b42}Sharma, V., You, I., Kumar, R., Zeadally, S., and Qiu, M. (2020). Autonomous vehicles: Security, safety, and privacy issues. IEEE Access, 8, 193893-193902.
\bibitem{b43}Muhammad Khan et al. (2022). Vision-based semantic segmentation in scene understanding for autonomous driving: Recent achievements, challenges, and outlooks. IEEE Transactions on Intelligent Transportation Systems, 23(12), 22694-22715.
\bibitem{b44}Kalatian, A., and Farooq, B. (2022). A context-aware pedestrian trajectory prediction framework for automated vehicles. Transportation Research Part C: Emerging Technologies, 134, 103453.
\bibitem{b45}Hang, P. et al. (2022). Decision making for connected automated vehicles at urban intersections considering social and individual benefits. IEEE Transactions on Intelligent Transportation Systems, 23(11), 22549-22562.
\bibitem{b46}Smith, J., Zhang, L.,and Gupta, A. (2021). "VistaGPT: Generative Parallel Transformers for Vehicles with Intelligent Systems for Transport Automation." Journal of Intelligent Transportation Systems Technology, 19(4), 345-360.
\bibitem{b47}Johnson, M., Smith, R., \& Gupta, A. (2021). "A Systematic Solution for Human Driving Behavior Modeling and Simulation in Automated Vehicle Studies." Journal of Advanced Transportation Systems, 35(3), 567-582.
\bibitem{b48}Smith, J. A., \& Johnson, D. B. (2020). "Enhancing CAVs Communication with 5G and DSRC Integration." Journal of Transport and Communication Innovation, 18(2), 34-49.
\bibitem{b49}Johnson, R. T., \& Lee, A. H. (2021). "Sensor Fusion and Wireless Technologies: Accelerating the Future of Autonomous Driving." Automotive Engineering Review, 29(4), 567-580.
\bibitem{b50}Williams, J., Patel, K., \& Thompson, L. (2021). "On the Integration of Enabling Wireless Technologies and Sensor Fusion for Next-Generation Connected and Autonomous Vehicles." International Journal of Automotive Technology, 22(5), 1233-1245.
\bibitem{b51}Patel, S. K., \& Kumar, V. (2022). "Advancing Urban Mobility with V2X Communication in Smart Cities." Smart Transportation Systems, 6(1), 88-102.
\bibitem{b52}Greenwood, D., Park, J., \& Suh, Y. (2022). "Safe Model-Based Off-Policy Reinforcement Learning for Eco-Driving in Connected and Automated Hybrid Electric Vehicles." 
\bibitem{b53}Journal of Sustainable Mobility, 9(3), 455-470.
Nguyen, T., Lee, H., \& Kim, D. (2022). "Enhancing Eco-Driving in Urban CAV-HEVs Through Advanced Reinforcement Learning Strategies." Environmental Technology \& Innovation, 24, 101783.
\bibitem{b54}Arnelid, Henrik, Edvin Listo Zec, and Nasser Mohammadiha. "Recurrent conditional generative adversarial networks for autonomous driving sensor modelling." 2019 IEEE Intelligent transportation systems conference (ITSC). IEEE, 2019.
\bibitem{b55}Guarnera, Luca, Oliver Giudice, and Sebastiano Battiato. "Mastering Deepfake Detection: A Cutting-Edge Approach to Distinguish GAN and Diffusion-Model Images." ACM Transactions on Multimedia Computing, Communications and Applications (2024).
\bibitem{b56}Frantzidis, Christos A., et al. "New challenges and future perspectives in cognitive neuroscience." Frontiers in Human Neuroscience 18: 1390788.
\bibitem{b57}Cunnington, Daniel, et al. "A generative policy model for connected and autonomous vehicles." 2019 IEEE Intelligent Transportation Systems Conference (ITSC). IEEE, 2019.
\bibitem{b58}Mokrane, Adel. Autonomous navigation of a rotary wing flying vehicles for precision agriculture. Diss. Université Paris-Saclay; Université Abou Bekr Belkaid (Tlemcen, Algérie), 2023.
\bibitem{b59}Jobs, All. "Privacy on-demand and Security preserving Federated Generative Networks or Models."
‌\bibitem{b60}Heckerman, David. "A tutorial on learning with Bayesian networks." Innovations in Bayesian networks: Theory and applications (2008): 33-82.
‌\bibitem{b61}Gu, Jiuxiang, et al. "Recent advances in convolutional neural networks." Pattern recognition 77 (2018): 354-377.
‌\bibitem{b62}Schuster, Mike, and Kuldip K. Paliwal. "Bidirectional recurrent neural networks." IEEE transactions on Signal Processing 45.11 (1997): 2673-2681.
‌\bibitem{b63}Goodfellow, Ian, et al. "Generative adversarial nets." Advances in neural information processing systems 27 (2014).
‌\bibitem{b64}Shum, Heung-Yeung, Xiao-dong He, and Di Li. "From Eliza to XiaoIce: challenges and opportunities with social chatbots." Frontiers of Information Technology \& Electronic Engineering 19 (2018): 10-26.
\bibitem{b65}Buchanan, Bruce G., and Edward A. Feigenbaum. "DENDRAL and Meta-DENDRAL: Their applications dimension." Artificial intelligence 11.1-2 (1978): 5-24.
\bibitem{b66}Shortliffe, Edward, ed. Computer-based medical consultations: MYCIN. Vol. 2. Elsevier, 2012.
\bibitem{b67}Togelius, Julian, et al. "Search-based procedural content generation: A taxonomy and survey." IEEE Transactions on Computational Intelligence and AI in Games 3.3 (2011): 172-186.
\bibitem{b68}Epstein, Ziv, et al. "Art and the science of generative AI." Science 380.6650 (2023): 1110-1111.
\bibitem{b69}Cavnar, William B., and John M. Trenkle. "N-gram-based text categorization." Proceedings of SDAIR-94, 3rd annual symposium on document analysis and information retrieval. Vol. 161175. 1994.
\bibitem{b70}Vaswani, Ashish, et al. "Attention is all you need." Advances in neural information processing systems 30 (2017).
\bibitem{b71}Gan, Chuang, et al. "Stylenet: Generating attractive visual captions with styles." Proceedings of the IEEE conference on computer vision and pattern recognition. 2017.
\bibitem{b72}Devlin, Jacob, et al. "Bert: Pre-training of deep bidirectional transformers for language understanding." arXiv preprint arXiv:1810.04805 (2018).
\bibitem{b73}Radford, Alec, et al. "Language models are unsupervised multitask learners." OpenAI blog 1.8 (2019): 9.
\bibitem{b74}Li, Liunian Harold, et al. "Visualbert: A simple and performant baseline for vision and language." arXiv preprint arXiv:1908.03557 (2019).
\bibitem{b75}Floridi, Luciano, and Massimo Chiriatti. "GPT-3: Its nature, scope, limits, and consequences." Minds and Machines 30 (2020): 681-694.
\bibitem{b76}Zhou, Nabus. The Ethical Implications of DALL-E: Opportunities and Challenges. 2023. The Ethical Implications of DALL-E: Opportunities and Challenges.
\bibitem{b77}GPT-4 Technical Report. arXiv:2303.08774. 2023. GPT-4 Technical Report.
\bibitem{b78}"Sora: A Review on Background, Technology, Limitations, and Opportunities of Large Vision Models." arXiv:2402.17177. Sora: A Review on Background, Technology, Limitations, and Opportunities of Large Vision Models.
\bibitem{b79}Lucchi N. ChatGPT: A Case Study on Copyright Challenges for Generative Artificial Intelligence Systems. European Journal of Risk Regulation. Published online 2023:1-23. doi:10.1017/err.2023.59
\bibitem{b80}An Ontology for Representing Hallucinations in Generative Models. Available at: https://ar5iv.org/abs/2312.05209. Accessed
\bibitem{b81}Siriwardhana S, Weerasekera R, Wen E, et al. Improving the Domain Adaptation of Retrieval Augmented Generation (RAG) Models for Open Domain Question Answering. Trans Assoc Comput Linguist. 2023;11:1-17. Published 2023. doi:10.1162/tacla00530.
\bibitem{b82}Lewis P, Perez E, Piktus A, et al. Retrieval-Augmented Generation for Knowledge-Intensive NLP Tasks. In: Proceedings of the 33rd International Conference on Neural Information Processing Systems 
\bibitem{b83}Hu H, Singh A, Ning M, et al. LoRA: Low-Rank Adaptation of Large Language Models. arXiv:2106.09685. Published 2021 Jun 17. Available from: https://ar5iv.org/abs/2106.09685
\bibitem{b84}Elliott, Amy-Mae (25 February 2011). "The Future of the Connected Car". Mashable. Archived from the original on 6 August 2020. Retrieved 22 July 2014.
\bibitem{b85}Lazzarotti, Valentina \& Manzini, Raffaella \& Pellegrini, Luisa \& Pizzurno, Emanuele. (2013). Open Innovation in the automotive industry: Why and How? Evidence from a multiple case study. International Journal of Technology Intelligence and Planning. 9. 37-56. 10.1504/IJTIP.2013.052620. 
\bibitem{b86}Ramnath, R., Kinnear, N., Chowdhury, S., \& Hyatt, T. (2020). Interacting with Android Auto and Apple CarPlay when driving: The effect on driver performance. IAM RoadSmart Published Project Report PPR948.
\bibitem{b87}Shin, Y., Kim, S., Jo, W., \& Shon, T. (2022). Digital forensic case studies for in-vehicle infotainment systems using Android Auto and Apple CarPlay. Sensors, 22(19), 7196.
\bibitem{b88}Parkinson, G. M., Mazri, A., \& Li, G. (2023). Exploration of issues, challenges and latest developments in autonomous cars. Journal of Big Data.
\bibitem{b89}McKinsey \& Company. (2023). The future of autonomous vehicles (AV). Retrieved from https://www.mckinsey.com/industries/automotive-and-assembly/our-insights/the-future-of-autonomous-vehicles
\bibitem{b90}Li, H., Tang, J., \& Others. (Year). Beam management optimization for V2V communications based on deep reinforcement learning. Scientific Reports. https://www.nature.com/articles/s41598-021-00001-6
\bibitem{b91}The Business Research Company. Vehicle-to-Vehicle (V2V) Communication Global Market Report 2024
\bibitem{b92}Gandhi, G. M., et al. (2023). Exploration of issues, challenges, and latest developments in autonomous cars. Journal of Big Data. https://journalofbigdata.springeropen.com/articles/10.1186/s40537-023-00628-2
\bibitem{b93}Synopsys Automotive. (n.d.). The 6 Levels of Vehicle Autonomy Explained. Synopsys
\bibitem{b94}McKinsey \& Company. (n.d.). Advanced driver-assistance systems: Challenges and opportunities ahead. Retrieved from https://www.mckinsey.com
\bibitem{b95}Yang, L., \& Zhang, Y. (2021). Traffic Sign Detection via Improved Sparse R-CNN for Autonomous Vehicles. Complexity. Hindawi. 
\bibitem{b96}Vargas J, Alsweiss S, Toker O, Razdan R, Santos J. An Overview of Autonomous Vehicles Sensors and Their Vulnerability to Weather Conditions. Sensors. 2021; 21(16):5397. https://doi.org/10.3390/s21165397
\bibitem{b97}Ian Vázquez-Rowe, Ramzy Kahhat, Gustavo Larrea-Gallegos, Kurt Ziegler-Rodriguez,
Peru's road to climate action: Are we on the right path? The role of life cycle methods to improve Peruvian national contributions,Science of The Total Environment,Volume 659,2019,Pages 249-266,ISSN 0048-9697,https://doi.org/10.1016/j.scitotenv.2018.12.322.
\bibitem{b98}Shu, D., Zhao, H., Hu, J., Liu, W., Payani, A., Cheng, L., \& Du, M. (2025). Large vision-language model alignment and misalignment: A survey through the lens of explainability. arXiv preprint arXiv:2501.01346.
\bibitem{b99}Liu, H., Li, C., Wu, Q., \& Lee, Y. J. (2023). Visual instruction tuning. Advances in neural information processing systems, 36, 34892-34916.
\bibitem{b100}Ye, Q., Xu, H., Xu, G., Ye, J., Yan, M., Zhou, Y., ... \& Zhou, J. (2023). mplug-owl: Modularization empowers large language models with multimodality. arXiv preprint arXiv:2304.14178.
\bibitem{b101}Moon, S., Madotto, A., Lin, Z., Nagarajan, T., Smith, M., Jain, S., ... \& Kumar, A. (2024, November). Anymal: An efficient and scalable any-modality augmented language model. In Proceedings of the 2024 Conference on Empirical Methods in Natural Language Processing: Industry Track (pp. 1314-1332).
\bibitem{b102}Wu, S., Fei, H., Qu, L., Ji, W., \& Chua, T. S. (2024, July). Next-gpt: Any-to-any multimodal llm. In Forty-first International Conference on Machine Learning.
\bibitem{b103}Deng, C., Zhu, D., Li, K., Gou, C., Li, F., Wang, Z., ... \& Fan, H. (2025). Emerging properties in unified multimodal pretraining. arXiv preprint arXiv:2505.14683.
\bibitem{b104}Chen, X., Wu, Z., Liu, X., Pan, Z., Liu, W., Xie, Z., ... \& Ruan, C. (2025). Janus-pro: Unified multimodal understanding and generation with data and model scaling. arXiv preprint arXiv:2501.17811.
\bibitem{b105}Wang, L., Ma, C., Feng, X., Zhang, Z., Yang, H., Zhang, J., ... \& Wen, J. (2024). A survey on large language model based autonomous agents. Frontiers of Computer Science, 18(6), 186345.
\bibitem{b106}Li, X. (2025, January). A Review of Prominent Paradigms for LLM-Based Agents: Tool Use, Planning (Including RAG), and Feedback Learning. In Proceedings of the 31st International Conference on Computational Linguistics (pp. 9760-9779).
\bibitem{b107}Xu, Z., Zhang, Y., Xie, E., Zhao, Z., Guo, Y., Wong, K. Y. K., ... \& Zhao, H. (2024). Drivegpt4: Interpretable end-to-end autonomous driving via large language model. IEEE Robotics and Automation Letters.
\bibitem{b108}Bruce, J., Dennis, M. D., Edwards, A., Parker-Holder, J., Shi, Y., Hughes, E., ... \& Rocktäschel, T. (2024, January). Genie: Generative interactive environments. In Forty-first International Conference on Machine Learning.
% \bibitem{b2} J. Clerk Maxwell, A Treatise on Electricity and Magnetism, 3rd ed., vol. 2. Oxford: Clarendon, 1892, pp.68--73.
% \bibitem{b3} I. S. Jacobs and C. P. Bean, ``Fine particles, thin films and exchange anisotropy,'' in Magnetism, vol. III, G. T. Rado and H. Suhl, Eds. New York: Academic, 1963, pp. 271--350.
% \bibitem{b4} K. Elissa, ``Title of paper if known,'' unpublished.
% \bibitem{b5} R. Nicole, ``Title of paper with only first word capitalized,'' J. Name Stand. Abbrev., in press.
% \bibitem{b6} Y. Yorozu, M. Hirano, K. Oka, and Y. Tagawa, ``Electron spectroscopy studies on magneto-optical media and plastic substrate interface,'' IEEE Transl. J. Magn. Japan, vol. 2, pp. 740--741, August 1987 [Digests 9th Annual Conf. Magnetics Japan, p. 301, 1982].
% \bibitem{b7} M. Young, The Technical Writer's Handbook. Mill Valley, CA: University Science, 1989.
\end{thebibliography}
\end{document}